\documentclass{article}

\PassOptionsToPackage{square,comma,numbers,sort&compress}{natbib}
\usepackage[preprint]{neurips_2022}

\usepackage[utf8]{inputenc} % allow utf-8 input
\usepackage[T1]{fontenc}    % use 8-bit T1 fonts
\usepackage{etoolbox}       % for possesive cite
\usepackage{url}            % simple URL typesetting
\usepackage{booktabs, tabularx}       % professional-quality tables
\usepackage{amsfonts}       % b lackboard math symbols
\usepackage{nicefrac}       % compact symbols for 1/2, etc.
\usepackage{microtype}      % microtypography
\usepackage{xcolor}         % colors
\usepackage[font=small,labelfont=bf]{caption}
\usepackage{amsfonts}       % blackboard math
\usepackage{nicefrac}       % compact symbols for 1/2, etc.
\usepackage{graphicx}
\usepackage{amsmath}
\allowdisplaybreaks
\usepackage{tabu}
\usepackage{amssymb}
\usepackage{gensymb}
\usepackage{mathtools}
\usepackage{wrapfig}
\usepackage{xcolor}
\usepackage{enumitem}
\usepackage{siunitx}
\usepackage{bm}
\usepackage{algorithm,algcompatible} % algorithms

\algnewcommand\INPUT{\item[\textbf{Input:}]}%
\algnewcommand\OUTPUT{\item[\textbf{Output:}]}%
\algnewcommand\algorithmicforeach{\textbf{for each}}
\algdef{S}[FOR]{ForEach}[1]{\algorithmicforeach\ #1\ \algorithmicdo} %foreach

\usepackage{hyperref}       % hyperlinks
\hypersetup{breaklinks=true,letterpaper=true,colorlinks}
\usepackage{cleveref}

\newcommand{\tobs}{\ensuremath{\bm{t}_\mathrm{obs}}}
\newcommand{\tfut}{\ensuremath{\bm{t}_\mathrm{fut}}}

\makeatletter

\patchcmd{\NAT@test}{\else \NAT@nm}{\else \NAT@nmfmt{\NAT@nm}}{}{}

\DeclareRobustCommand\citepos
  {\begingroup
   \let\NAT@nmfmt\NAT@posfmt% ...except with a different name format
   \NAT@swafalse\let\NAT@ctype\z@\NAT@partrue
   \@ifstar{\NAT@fulltrue\NAT@citetp}{\NAT@fullfalse\NAT@citetp}}

\let\NAT@orig@nmfmt\NAT@nmfmt
\def\NAT@posfmt#1{\NAT@orig@nmfmt{#1's}}

\makeatother
\title{Why Did This Model Forecast This Future?\\ Closed-Form Temporal Saliency Towards Causal Explanations of Probabilistic Forecasts}
\author{Chirag Raman\qquad\qquad Hayley Hung\qquad\qquad Marco Loog\\ \\
Delft University of Technology, Delft, The Netherlands\\
\texttt{\{c.a.raman, h.hung, m.loog\}@tudelft.nl}
}

\begin{document}

\maketitle

\begin{abstract}
    Forecasting tasks surrounding the dynamics of low-level human behavior are of significance to multiple research domains. In such settings, methods for explaining specific forecasts can enable domain experts to gain insights into the predictive relationships between behaviors. In this work, we introduce and address the following question: given a probabilistic forecasting model how can we identify observed windows that the model considers salient when making its forecasts? We build upon a general definition of information-theoretic saliency grounded in human perception and extend it to forecasting settings by leveraging a crucial attribute of the domain: a single observation can result in multiple valid futures. We propose to express the saliency of an observed window in terms of the differential entropy of the resulting predicted future distribution. In contrast to existing methods that either require explicit training of the saliency mechanism or access to the internal states of the forecasting model, we obtain a closed-form solution for the saliency map for commonly used density functions in probabilistic forecasting. We empirically demonstrate how our framework can recover salient observed windows from head pose features for the sample task of speaking-turn forecasting using a synthesized conversation dataset.
\end{abstract}

% 1. Situate the work. Why is this important?
% 2. Proposal, contributions
% 3. Derivation, experiments
% 4. Results
% 5. Take away message

\section{Introduction}
\label{sec:introduction}

The existence of multiple valid futures for a given observed sequence is a crucial attribute of several forecasting tasks, especially surrounding the dynamics of low-level human behavior. These tasks include the forecasting of trajectories of pedestrians \cite{salzmannTrajectronDynamicallyFeasibleTrajectory2021, rudenko2020human, huangSTGATModelingSpatialTemporal2019, zhangSRLSTMStateRefinement2019, mohamedSocialSTGCNNSocialSpatioTemporal2020}, vehicles \cite{gillesTHOMASTrajectoryHeatmap2022,zhaoTNTTargetdriveNTrajectory2020, carrascoSCOUTSociallyCOnsistentUndersTandable2021, zengDSDNetDeepStructured2020}, and autonomous robots \cite{vemula2017modeling, ivanovic2021propagating}, or other more general nonverbal cues of humans \cite{ramanSocialProcessesSelfSupervised2021, adeliSociallyContextuallyAware2020, nguyenContextAwareHumanBehaviour2022, barqueroDidnSeeThat2022, yaoMultipleGranularityGroup2018a} and artificial virtual agents \cite{ahujaReactNotReact2019} in group conversation settings.  Consequently, rather than making point predictions, several machine learning methods in these settings have attempted to forecast a distribution over plausible futures \cite{mohamedSocialSTGCNNSocialSpatioTemporal2020, ramanSocialProcessesSelfSupervised2021}. In this work, we introduce and address a specific research question towards gaining domain-relevant insights into such forecasts: how can we identify preceding observed sequences that are salient for a given model towards making \textit{probabilistic} forecasts over a particular future window? 

Similar questions have been recently posed for models making point forecasts across various application domains \cite{oreshkinNBEATSNeuralBasis2020a, limTemporalFusionTransformers2020a, panTwoBirdsOne2021}. The broader goal here has been on designing architectures that learn to produce forecasts that are not only accurate, but also interpretable. However, in what \citet{liptonMythosModelInterpretability2017} terms \textit{The Mythos of Interpretability}, the notion of what renders these models interpretable largely has non-overlapping motivations and attributes across different works. Examples of these notions of interpretability\footnote{\citet{BARREDOARRIETA202082} argue for the importance of distinguishing between interpretability and explainability as different concepts, the latter denoting any active action or procedure taken by a model with the intent of clarifying or detailing its internal functions. From this perspective, what the cited works term interpretability is closer to the notion of explainability. Also see \cite{BARREDOARRIETA202082} for an overview of taxonomies of explainability approaches beyond time-series forecasting.} and the associated approaches for forecasting tasks include: (i) injecting a suitable inductive bias into the model through a set of basis functions and identifying how they combine to produce an output \cite{oreshkinNBEATSNeuralBasis2020a} (an approach that has recently been applied to the probabilistic setting  as well \cite{rugamerProbabilisticTimeSeries2022}); (ii) employing a self-attention mechanism to learn temporal patterns while attending to a common set of features \cite{limTemporalFusionTransformers2020a}; and (iii) applying the notion of saliency maps from computer vision \cite{dabkowski2017real} to time-series data as a measure of how much each feature contributes to the final forecast \cite{panTwoBirdsOne2021}. Irrespective of the notion of interpretability, these methodologies are underpinned by two common attributes: (a) the interpretability mechanism needs explicit training as part of the model architecture, and (b) what constitutes a \textit{good} explanation is subject to the biases, intuition, or the visual assessment of the human observer \cite{millerExplanationArtificialIntelligence2019, adebayo2018sanity}; a phenomenon we refer to as the interpretation lying in the eye of the beholder. This is especially true for when saliency maps have been used as tools for post-hoc explanations in computer vision: the computed map may not measure the intended saliency, and even be independent of both the model and data generating process \citet{adebayo2018sanity}. Indeed, things are further confounded by the lack of a common notion of saliency. As \citeauthor{BARREDOARRIETA202082} point out, ``there is absolutely no consistency behind what is known as saliency maps, salient masks, heatmaps, neuron activations, attribution, and other approaches alike'' \citep[Sec.~5.3]{BARREDOARRIETA202082}. 

\subsection{Revisiting Explanations, and Drawbacks of Existing Methods}
Against the backdrop of this research landscape, we begin by revisiting the question: what constitutes an explanation? \citet{millerExplanationArtificialIntelligence2019} recently turned to the vast body of research in philosophy, pscyhology, and cognitive science on the topic and highlighted the importance of causality in explanation. Specifically, in Table~\ref{tab:explanations} we reproduce the simple categorization of explanatory questions he proposed based on Pearl and Mackenzie's \textit{Ladder of Causation} \cite{pearl2018book}. Within this categorization, Miller argues that the why-questions are the most challenging, because they use the most sophisticated reasoning and are contrastive in nature. 

Most of the existing interpretability approaches (involving injecting inductive biases \cite{oreshkinNBEATSNeuralBasis2020a, rugamerProbabilisticTimeSeries2022} and attention-based mechanisms \cite{limTemporalFusionTransformers2020a}) can be considered to be associative within Miller's categorization, since they reason about the (unobserved) importance of input features conditioned on one realization of the future prediction (the event). On the surface, the perturbation-based saliency methods for interpretable forecasting can be considered to be counterfactual in approach: \citet{panTwoBirdsOne2021} learn a saliency mask that perturbs different parts of the input (simulating alternative `causes') in a way that minimizes an error metric between the corresponding prediction and ground-truth future (the `event'). So while saliency-based approaches can be promising in answering such why-questions, we identify several concerns in applying such an approach directly for causal reasoning: (i) the alternative `causes' generated by noisy perturbations to parts the input may not correspond to valid real behaviors; (ii) the notion of saliency in terms of minimizing an error metric is an arbitrary choice, and requires the ground-truth future for optimizing the saliency mask for a specific instance; (iii) the saliency map identifies instance-specific attributes rather than aggregating temporal patterns across all instances; and (iv) the map needs to be explicitly trained (for every instance in the case of \cite{panTwoBirdsOne2021}).  
\begin{table}[t]
\caption{Classes of Explanatory Question and the Reasoning Required to Answer. Reproduced from \citet[Table~3]{millerExplanationArtificialIntelligence2019}} \label{tab:explanations}
    % \centering % not required, text width
    \begin{tabularx}{\textwidth}{llX}
        \toprule
        \textbf{Question} & \textbf{Reasoning} & \textbf{Description}\\
        \midrule
         What? & Associative & Reason about which unobserved events could have occurred given the observed events\\\\
         How? & Interventionist & Simulate a change in the situation to see if the event still happens\\\\
         Why? & Counterfactual & Simulating alternative causes to see whether the event still happens\\
        \bottomrule
    \end{tabularx}
    \vspace{-10pt}
\end{table}

\subsection{Our Approach: Building upon a Unifying Framework of Bottom-Up Saliency}
In this work, our central idea for temporal saliency in forecasting settings is as follows: the saliency of a preceding observed window is related to the \textit{uncertainty over the future window of interest resulting from its observation}. Our framework addresses all the aforementioned concerns we identified with saliency-based approaches in forecasting as follows: (i) the alternate `causes' in our framework pertain to real preceding windows of actual observed behavior; (ii) our information-theoretic perspective builds upon a fundamental definition of saliency grounded in bottom-up preattentive human perception \cite{loogInformationTheoreticPreattentive2011, vanderheijden1996perception}, and can be used to compute saliency for unseen test data where the ground-truth future is unavailable; (iii) we utilize a distribution over the futures that the model believes could have occurred over a future window rather than the single future that did occur: what a model considers plausible for a single forecast encapsulates the temporal patterns of behavior it has learned from the entire dataset; and (iv) we derive a closed-form expression for the saliency map that can be applied to any model that outputs a probability distribution for its forecasts without additional training or optimization. 

It is reasonable to assume that a model that accurately estimates a distribution over plausible futures for an input has captured structural predictive dependencies between features in the data. Our long-term goal is to enable causal insights using such forecasts. However, the current state of explainable artificial intelligence is characterized by a lack of consensus over whether explanations are truly meaningful for real-world applications. Within this research landscape, rather than directly discovering the behavioral causes of a forecast, we view the present work as part of a prudent two-step methodology that incorporates a domain-expert in the loop. First, our proposed framework here can be used to identify \textit{windows} salient towards a model's forecasts as candidate causes. Second, these candidate causes can be presented to the domain-expert for subsequent dedicated analysis to investigate whether the salient relationships between the  \textit{behaviors} within the observed and future windows are truly causal.

The rest of this paper is organized as follows: in Section~\ref{sec:background} we establish the general framework of bottom-up preattentive saliency which we build upon. In Section~\ref{sec:method} we formalize our proposed framework for temporal saliency in forecasting settings, and demonstrate its empirical working in Section~\ref{sec:illustration}. We review related literature in Section~\ref{sec:related-work} and conclude with a discussion in Section~\ref{sec:discussion}.

\section{Background: Information Theoretic Preattentive Saliency in Images}
\label{sec:background}

\citet{loogInformationTheoreticPreattentive2011} developed a general closed-form expression for saliency using a surprisal-based operational definition of bottom-up attention from the field of computational visual perception. Let $L:\mathbb{R}^n \to \mathbb{R}^d$ be a general $n$-dimensional $d$-valued image. A continuously differentiable feature mapping $\phi: \mathbb{R}^n \to \mathbb{R}^N$ relates every image location $x \in \mathbb{R}^n$ from $L$ to $N$ features. Further, let $p_X$ denote the probability density function over image locations $x$ and $p_\Phi$ denote the probability density function over all feature vectors $\phi(x)$. The distribution $p_X$ captures any prior knowledge that would make one location more salient than another. In the absence of such prior knowledge, $p_X$ is typically chosen to be uniform.  

The saliency $S(x)$ of a location $x$ is then defined in terms of the amount of information\textemdash or surprise\textemdash of its associated feature vector $\phi(x)$ relative to the other feature vectors extracted from the same image. The intuition is that the larger the information content of a certain combination of features, $-\log p_\Phi(\phi(x))$, the more salient the location $x$:
\begin{equation}
    S(x) > S(x') \iff -\log p_\Phi(\phi(x)) > -\log p_\Phi(\phi(x')).
\end{equation}
This general definition unifies seemingly different definitions of saliency encountered in the literature. It relates to salient observations being considered unexpected, rare, or surprising \cite{walker1998locating, garcia2001information, itti2001computational, torralba2003modeling}. It also incorporates the notion of saliency considered in decision-theoretic settings \cite{torralba2003modeling, lingyun2007information}, where bottom-up saliency increases with an increase in the information associated with feature vectors \cite{rosenholtz1999simple, gao2007bottom}.

Contrary to approaches that determine saliency maps through an explicit data-driven density estimation \cite{torralba2003modeling, bruce2005features, itti2005principled, gao2007bottom, li2013saliency, simonyan2013deep, li2016visual, zhou2016learning, dabkowski2017real, fong2017interpretable, jiang2019image, huang2020image}, a closed form expression for saliency can be given once the feature mapping $\phi$ is fixed. When $\phi$ is continuously differentiable, 
% and the feature dimension $N$ is greater than twice the spatial dimension $n$, it follows from a classical result on regular mappings by \citet{whitneyDifferentiableManifolds1936} that $\phi$ provides an embedding of $\mathbb{R}^n$ in $\mathbb{R}^N$. Stated differently, if sufficient features are chosen to describe every image location, they provide a bijective map between $\mathbb{R}^n$ and its image $\phi(\mathbb{R}^n) \subset \mathbb{R}^N$ under $\phi$.
% Following Whitney's general result, 
the information content $-\log p_{\Phi}$ on $\phi(\mathbb{R}^n) \subset \mathbb{R}^N$ over all image features can be obtained from $\log p_X$ through a simple change of variables \cite{boothby1975introduction} from $x$ to $\phi(x)$. The amount of information for the feature vector $\phi(x)$ at every location $x$, and thereby the saliency $S(x)$, is then given by the expression:
\begin{align}
    -\log  p_{\Phi}(\phi(x)) &= -\log \frac{p_X(x)}{\sqrt{\det(J^t_{\phi}(x)J_{\phi}(x))}}\\
                             &= -\log p_X(x) + \frac{1}{2} \log\det(J^t_{\phi}(x)J_{\phi}(x)), \label{eq:saliency}
\end{align}
where $J_\phi : \mathbb{R}^n\to\mathbb{R}^{N \times n}$ denotes the Jacobian matrix of $\phi$, and $\_^t$ indicates matrix transposition.

A crucial implication of this result is that the computation of saliency can be performed based on purely \textit{local} measurements and without the need to refer to previously observed data or any explicit density estimate of these. 
% Moreover, as Loog shows, the requirement $N > 2n$ is not necessary for $\phi$ to be injective, but allows for the easy demonstration of how a closed-form expression of saliency in Equation~\ref{eq:saliency} may be obtained; the expression remains a valid choice for defining a saliency map irrespective of the values of $n$ and $N$ \citep[see][Sec.~3]{loogInformationTheoreticPreattentive2011}.
Following Equation~\ref{eq:saliency}, \citeauthor{loogInformationTheoreticPreattentive2011} simplifies the definition of the saliency map to
\begin{equation}
    S(x) \coloneqq \det(J^t_{\phi}(x)J_{\phi}(x)), \label{eq:saliency-map}
\end{equation}
given that a monotonic transformation of the saliency map does not essentially alter the map.

\section{Closed-Form Temporal Saliency for Social Behavior Forecasting}
\label{sec:method}

\subsection{Setup and Notation}

Following the notation from the formulation of the task of Social Cue Forecasting (SCF) \cite{ramanSocialProcessesSelfSupervised2021}, let $\tobs~\coloneqq~[o1, o2, ..., oT]$ denote a window of consecutively increasing observed timesteps, and $\tfut~\coloneqq~[f1, f2, ..., fT]$ denote an unobserved future time window, with $f1>oT$. Note that
$\tfut$ and $\tobs$ can be of different lengths, and $\tfut$ need \textit{not} immediately follow $\tobs$. Given a set of $n$ interacting participants, let us denote their behavioral cues over $\tobs$ and $\tfut$ respectively as $\bm{X} \coloneqq [\bm{b}^i_{t}; t \in \tobs]_{i=1}^n$ and $\bm{Y} \coloneqq [\bm{b}^i_{t}; t \in \tfut]_{i=1}^n$. The vector $\bm{b}^i_{t}$ encapsulates the multimodal cues of interest from participant $i$ at time $t$, such as head and body pose, facial expressions, gestures, etc. The task of SCF is to forecast a distribution over possible future behavioral cues for a given observed sequence $\bm{X}$ of the same cues, denoted by the probability density function $p_{Y|X}$. Here $Y$ and $X$ denote the multivariate random variables associated with the future and observed sequences respectively. While the original formulation of SCF dealt with conversation settings, it is general enough to also apply to other social settings such as pedestrian trajectory forecasting, where the social cue is simply the location of an individual. In the rest of this work, for simplicity we denote the feature array at an individual timestep $t$ in $\tobs$ and $\tfut$ as $\bm{X}_t$ and $\bm{Y}_t$ respectively, even though they correspond to the same behavioral cues. The focus of this work is to compute the saliency $S(\tobs)$ of an observed $\tobs$ towards the future occurring over a fixed choice of $\tfut$.

\subsection{Defining \boldmath{\texorpdfstring{$\phi$}{phi}} in Terms of the Uncertainty over the Future Window \boldmath{\texorpdfstring{$\tfut$}{tfut}}}
The closed-form expression for information-based spatial saliency in Equation~\ref{eq:saliency} makes it explicit that the choice of the feature mapping $\phi$ determines the actual form of the saliency map. 
To extend the saliency framework to forecasting settings, we begin with the following intuition: the saliency of an observed window $\tobs$ is directly related to how certain or uncertain it makes the resulting future over $\tfut$. This idea is related to the notion of surprisal; an observation that changes the certainty of the future is surprising, and thereby salient.
We formalize this intuition by mapping the window $\tobs$ to the differential entropy of the predicted future distribution over $\tfut$, conditioned on the observed features $\bm{X}$.  That is, we define $\phi : \tobs \mapsto h(Y|X=\bm{X})$, where the conditional differential entropy of $Y$ given $\{X=\bm{X}\}$ is defined as
\begin{equation}
    h(Y|X=\bm{X}) \coloneqq - \int p_{Y|X}(\bm{Y}|\bm{X})\log p_{Y|X}(\bm{Y}|\bm{X})d\bm{Y}. \label{eq:entropy-def}
\end{equation}
Following Equation~\ref{eq:entropy-def}, our framework for computing the saliency map is as follows. First we identify a specific $\tfut$ of interest. This could correspond to a semantic event of forecasting interest into which a domain-expert seeks to get some insight, such as a speaking-turn change \cite{keitel2015use, levinsonTimingTurntakingIts2015} an interaction termination \cite{bohusManagingHumanRobotEngagement2014, vandoornRitualsLeavingPredictive2018}, or a synchronous behavior event \cite{bilakhiaAudiovisualDetectionBehavioural2013}. Looking back in time before $\tfut$, we then compute $h(Y|X=\bm{X})$ for different observed behaviors $\bm{X}$ corresponding to different locations of a sliding $\tobs$ within a look-back period determined to be reasonable for the specific application. The computed differential entropy values are then inserted into Equation~\ref{eq:saliency-map} to obtain the saliency of the different behaviors over the $\tobs$ locations towards forecasting the future over the chosen $\tfut$. The general approach is summarized in Algorithm~\ref{alg:framework}.

\begin{algorithm}[t] 
    \caption{Temporal Saliency in Probabilistic Forecasting} \label{alg:framework}
    \begin{algorithmic}[1]
        \REQUIRE The probability density function $p_{Y|X}$
        \INPUT A fixed $\tfut$ of interest, a sequence of $m$ preceding observed windows $O = [\tobs^1,\ldots,\tobs^m]$, and the behavioral features $\bm{X}^j$ for every $\tobs^j$
        \OUTPUT The saliency map $S(O)$ over the observed windows
        \Statex
        \ForEach{$\tobs^j \in O$}
        \State Compute the feature mapping  $\phi(\tobs^j) \leftarrow h(Y|X=\bm{X}^j)$
        \ENDFOR
        \Statex
        \State Compute the saliency map $S(\tobs) \leftarrow \det(J^t_{\phi}(\tobs)J_{\phi}(\tobs))$
    \end{algorithmic}
\end{algorithm}

% \subsection{Favorable Properties of Entropy and Caveats}
Differential entropy possesses certain favorable properties that make it a suitable choice as $\phi$ for computing the saliency map in forecasting settings. To begin, the scale of the forecast density does not affect the resulting saliency map \cite[see][Theorem~8.6.4]{coverElementsInformationTheory}:
\begin{align}
    h(aY) &= h(Y) + \log|a|, \text{for } a \neq 0, \text{and}\\
    h(\bm{A}Y) &= h(Y) + \log|\det(\bm{A})|, \text{when $\bm{A}$ is a square matrix}.
\end{align}
That is, scaling the distribution changes the differential entropy by only a constant factor. So the saliency map resulting from inserting the entropy into Equation~\ref{eq:saliency-map} remains unaffected since the Jacobian term only depends on the relative change in entropy across different choices of $\tobs$. Similarly, translating the predicted density leaves the saliency map unaffected \cite[see][Theorem~8.6.3]{coverElementsInformationTheory}:
\begin{equation}
    h(Y+c) = h(Y).
\end{equation}

\subsection{Computing \texorpdfstring{$ h(Y|X=\bm{X})$}{h(Y|X=X)}} \label{subsec:computing-entropy}
To compute the differential entropy of the future distribution, we need the density function $p_{Y|X}$. Typically, this density is modeled as a multivariate Gaussian distribution \cite{ramanSocialProcessesSelfSupervised2021, NEURIPS2019_0b105cf1, salinasDeepARProbabilisticForecasting2019, NEURIPS2018_5cf68969}. 
When the decoding of the future is non-autoregressive, the parameters of the distributions for all $t \in \tfut$ are estimated in parallel. In these cases, the differential entropy has a closed-form expression. Assuming a $d$-dimensional predicted Gaussian distribution with mean $\bm{\mu}$ and covariance matrix $\bm{K}$, the expression for the entropy of a multivariate Gaussian distribution is given by \cite[see][Theorem~8.4.1]{coverElementsInformationTheory}
\begin{equation}
    h(Y|X=\bm{X}) = h(\mathcal{N}_d(\bm{\mu}, \bm{K})) = \frac{1}{2}\log [(2\pi e)^d \det(\bm{K})]. \label{eq:gaussian-entropy}
\end{equation}
When $\bm{K}$ is diagonal, so that the predicted distribution is factorized over participants and features, we can simply sum the $\log$ of the individual variances to obtain the feature mapping $\phi$. 
Note that from Equation~\ref{eq:gaussian-entropy}, for a multivariate Gaussian distribution, the differential entropy only depends on the covariance, or the \textit{spread} of the distribution, aligning with the notion of differential entropy as a measure of total uncertainty. (See \citetext{\citealp[Tab.~17.1]{coverElementsInformationTheory}; \citealp[]{lazo1978entropy}} for closed-form expressions for a large number of commonly employed probability density functions.)

\begin{wrapfigure}{r}{0.5\columnwidth}
    \vspace{-12pt}
    \centering 
    \includegraphics[width=0.49\columnwidth]{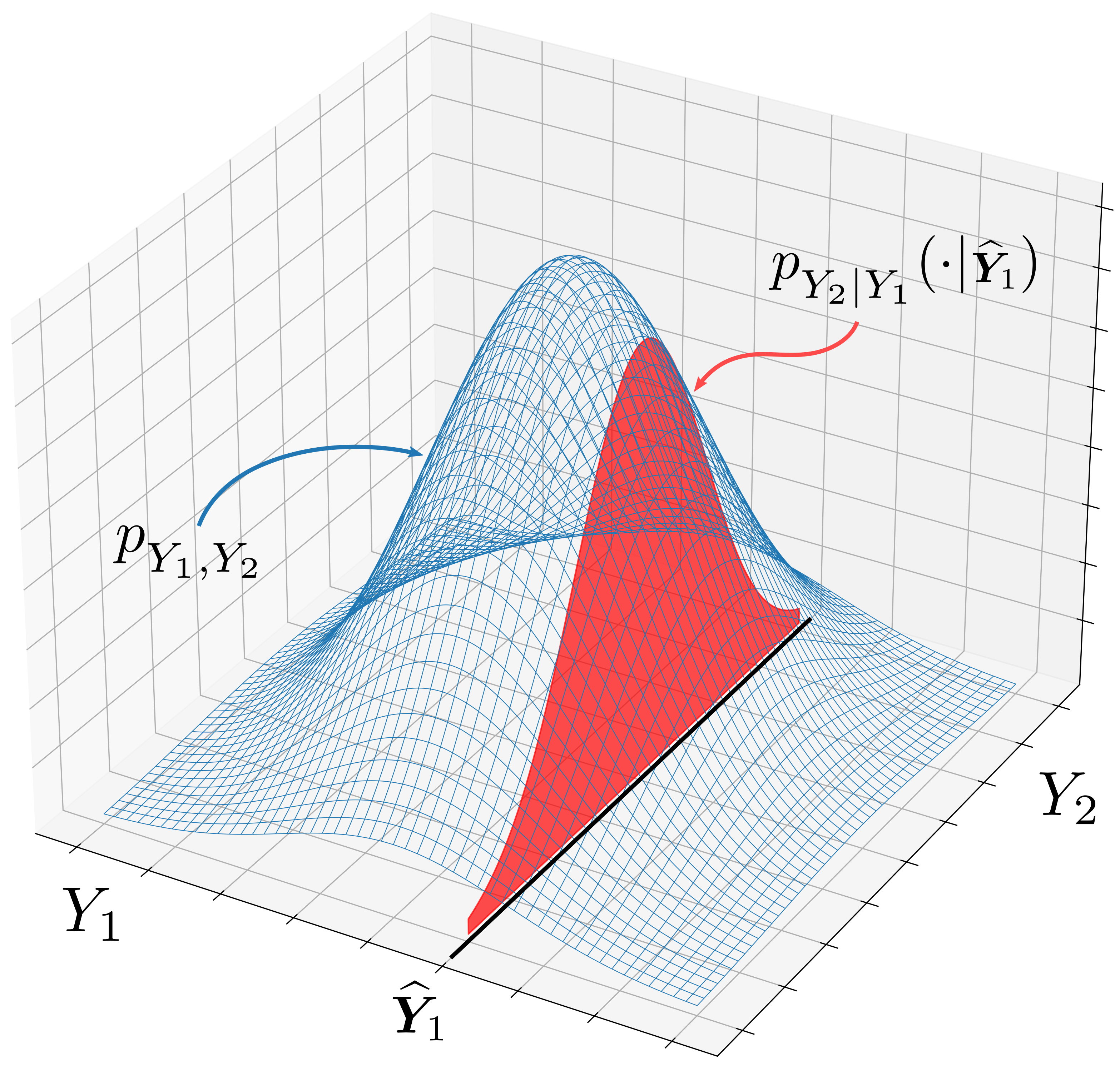}
    \caption{Illustrating predicted densities under the greedy autoregressive decoding approach for two timesteps. For simplicity, we depict a joint Gaussian distribution and omit the conditioning on $\bm{X}$ everywhere.}\label{fig:autoreg-densities}
\vspace{-10pt}
\end{wrapfigure}

Another common approach for inferring the future density function is the use of probabilistic autoregressive decoders \cite{ramanSocialProcessesSelfSupervised2021,salzmannTrajectronDynamicallyFeasibleTrajectory2021,haNeuralRepresentationSketch2017, salinasDeepARProbabilisticForecasting2019}.
Here, one possible decoding approach \cite{haNeuralRepresentationSketch2017, salzmannTrajectronDynamicallyFeasibleTrajectory2021} involves taking a specific sample $\widehat{\bm{Y}}_t$ from the density predicted at each $t \in \tfut$, and passing it back as input to the decoder for the estimation of the density at timestep $t+1$.
Therefore, the density at $t+1$ depends on the randomness introduced in sampling $\widehat{\bm{Y}}_t$.  
This yields an arbitrarily complex form for the joint distribution $p_{Y|X}$ (over all $t \in \tfut$).
\figurename~\ref{fig:autoreg-densities} illustrates the concept for two timesteps. Here, a single forecast would only output the shaded red distribution for $Y_2$. This precludes the direct computation of $h(Y_1,Y_2)$ that requires the full joint distribution $p_{Y_1,Y_2}$.
In such cases, we have two broad options: using a simplifying assumption to retain computational simplicity, or approximating the differential entropy by sampling.

The simpler option is to redefine our feature-mapping as $\phi : \tobs \mapsto \sum_{t \in \tfut} h(Y_t|\widehat{\bm{Y}}_{<t},\bm{X})$, i.e. we sum the differential entropy of the individual densities estimated at each timestep as an approximation of the total uncertainty over the predicted sequence. 
Note that following the chain rule for differential entropy (\cite[see][Eq.~8.62]{coverElementsInformationTheory}), the joint entropy can indeed be written as the sum of individual conditionals. However, in general,
\begin{equation}
    h(Y|X=\bm{X}) = \sum_{t \in \tfut}h(Y_t|Y_{<t},\bm{X}) \neq \sum_{t \in \tfut} h(Y_t|\widehat{\bm{Y}}_{<t},\bm{X}).
\end{equation}
Employing this approximation (of summing the entropies across all timesteps) relies on the observation that for autoregressive decoding, the parameters of the  predicted distribution for $Y_t$ is computed as a deterministic function of the decoder hidden state. That is, $Y_t$ is conditionally independent of $Y_{<t}$ given the hidden state of the decoder $\bm{s}_t$ at timestep $t$. 
The underlying assumption is that for a well-trained decoder, $\bm{s}_t$ encodes all relevant information from other timesteps to infer the distribution of $Y_t$. So at inference, despite being a function of the single sample $\widehat{\bm{Y}}_{t-1}$, the predicted distribution conditioned on $\bm{s}_t$ provides a reasonable estimate of the the uncertainty in $Y_t$. The benefit of employing this assumption is that when each $Y_t$ is modeled using a density function that has a closed-form expression for differential entropy \citetext{\citealp[Tab.~17.1]{coverElementsInformationTheory}; \citealp[]{lazo1978entropy}}, every item in the sum can be computed anlaytically, and we obtain a closed-form expression for the saliency map. Apart from a multivariate Gaussian, the other common choice for modeling $Y_t$ is using a Gaussian mixture \cite{haNeuralRepresentationSketch2017, salzmannTrajectronDynamicallyFeasibleTrajectory2021}. While a closed-form expression for the differential entropy of a Gaussian mixture is not known, approximations that approach the true differential entropy can be obtained efficiently \cite{huberEntropyApproximationGaussian2008, zhang2017approximating, michalowiczCalculationDifferentialEntropy2008} to directly compute the feature mapping $\phi$. 

What if the differential entropy for $Y_t$ does not have an analytical expression or a computationally efficient approximation? In such cases, $h(Y|X=\bm{X})$ can be estimated using sampling or other non-parametric approaches \cite{ariel2020estimating, brewer2017computing, ajglDifferentialEntropyEstimation2011, beirlant1997nonparametric}. While computationally more expensive than parametric methods, such sampling-based methods can provide approximations that converges to the true entropy.

Another practical consideration arises from the fact that we consider different candidate $\tobs$ to evaluate how their observation changes the uncertainty over a fixed $\tfut$. So $\tfut$ need not immediately follow $\tobs$. As \citet{ramanSocialProcessesSelfSupervised2021} point out, a model for forecasting human behavior ought to support such a requirement given that domain experts are often interested in behavioral phenomena that occur after an arbitrary offset \citep[see][Sec.5]{ramanSocialProcessesSelfSupervised2021}. In such cases, the typical approach which starts decoding immediately after the end of $\tobs$ would entail discarding the predictions in the gap between $\tobs$ and $\tfut$. This could cascade prediction errors and lead to worse estimates of $p_{Y|X}$. 
% We recommend that users of our framework in practical settings account for this potential source of prediction errors in the underlying model. 
One approach to avoid such errors in practice is for the underlying model to encode the offset $\Delta t = f1 - oT$ as an input and commence decoding at timestep $f1$ \cite{ramanSocialProcessesSelfSupervised2021}.  

\section{Empirical Demonstration Using Synthesized Variance over Futures}
\label{sec:illustration}
\begin{figure}[t!]
\makebox[\textwidth][c]{\includegraphics[width=\textwidth]{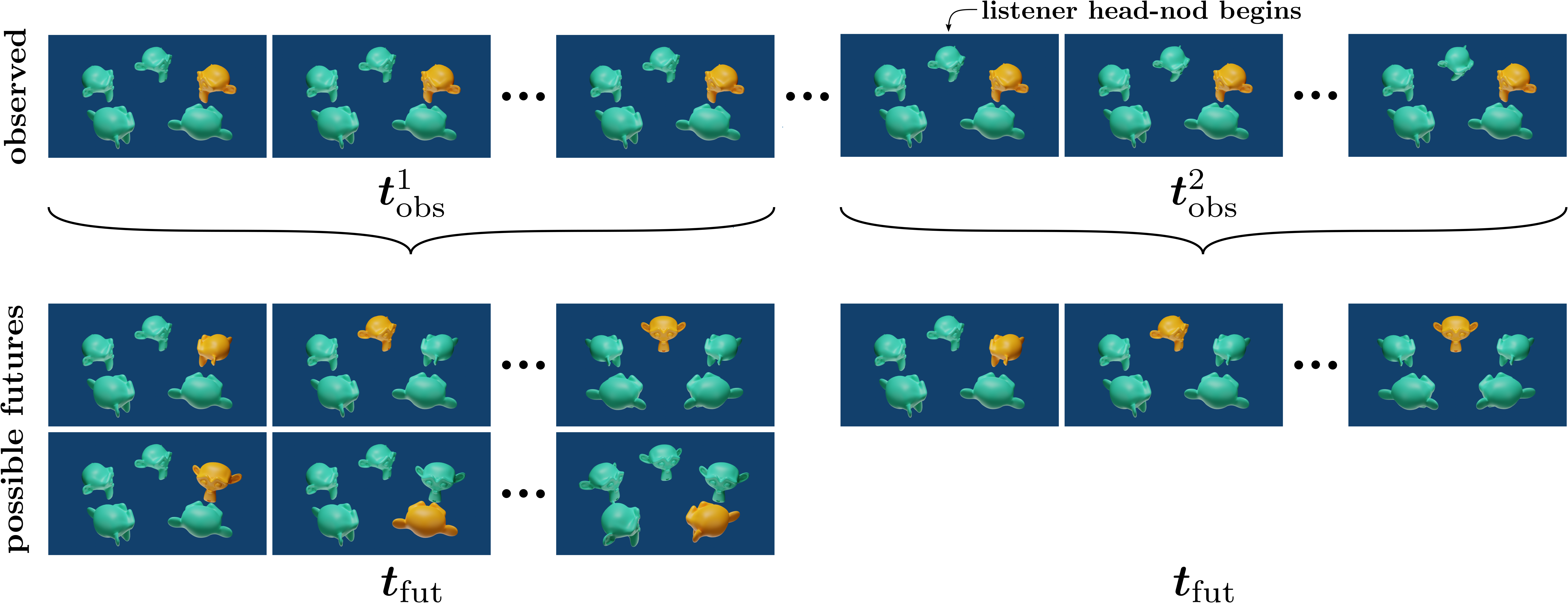}}
\caption{\textbf{Illustrating the synthetic conversation dynamics dataset.} Speakers are denoted in orange and listeners in green. For a fixed $\tfut$ we depict two preceding $\tobs$ windows. By construction, when observing a stable speaking turn over $\tobs^1$, two valid futures are possible over $\tfut$. These correspond to a turn handover to the immediate left or right of the current speaker. Over $\tobs^2$, when a listener nods to indicate the desire to take the floor, the future over $\tfut$ becomes certain, corresponding to the listener successfully taking over the speaking turn. Here $\tobs^2$ is consequently more salient than $\tobs^1$ towards forecasting the turn change over $\tfut$.     
}
\label{fig:dataset}
\vspace{-7pt}
\end{figure}

The best a well-trained model can do is to predict the true variance over plausible futures given an observation. In this section we use a synthetic dataset of simplified conversation dynamics to demonstrate how our framework empirically retrieves the windows that are salient by construction. Note that the model does not need to have any conception of saliency, and as such our framework doesn't involve any training using a top-down notion of saliency. The goal is to illustrate how our framework retrieves what would constitute as bottom-up salient to a well-trained forecasting model. 

\subsection{The Dataset}

For illustration, we choose a sample forecasting task that has received extensive domain interest over the last decade: forecasting a turn change in a multiparty conversation \cite{petukhova2009s, de2009multimodal, ishii2013predicting, rochet-capellanTakeBreathTake2014, keitel2015use, ishiiPredictionNextUtteranceTiming2017, malik2020speaks}. We synthesize a group conversation following established patterns of social behavior. First, the visual focus of attention of listeners is usually the speaker, while the speaker might look at different listeners over the speaking turn \cite{hungInvestigatingAutomaticDominance2008}. Second, head gestures and gaze patterns are predictive of the next speaker \cite{petukhova2009s, de2009multimodal, malik2020speaks, ishiiPredictionNextUtteranceTiming2017}. We implement a simplified version of these dynamics as follows. The speaker rotates towards the geometric center of the formation after acquiring a speaking turn, while the listeners orient towards the speaker. A listener nods their head to indicate a desire to acquire the floor, following which the current speaker rotates to look at the listener and hands over the speaking turn. We represent head pose with quaternions, given their common use in representing human motion and pose \cite{ramanSocialProcessesSelfSupervised2021, pavlloQuaterNetQuaternionbasedRecurrent2018}. We simulate the turn changes to occur once clockwise and once anticlockwise around the group so that each participant yields the floor once each to the participant on their immediate left and right. We provide the dataset and an animated visualization in the Suppementary material.

\begin{wrapfigure}{r}{0.55\columnwidth}
    \vspace{-14pt}
    \centering 
    \includegraphics[width=0.55\columnwidth]{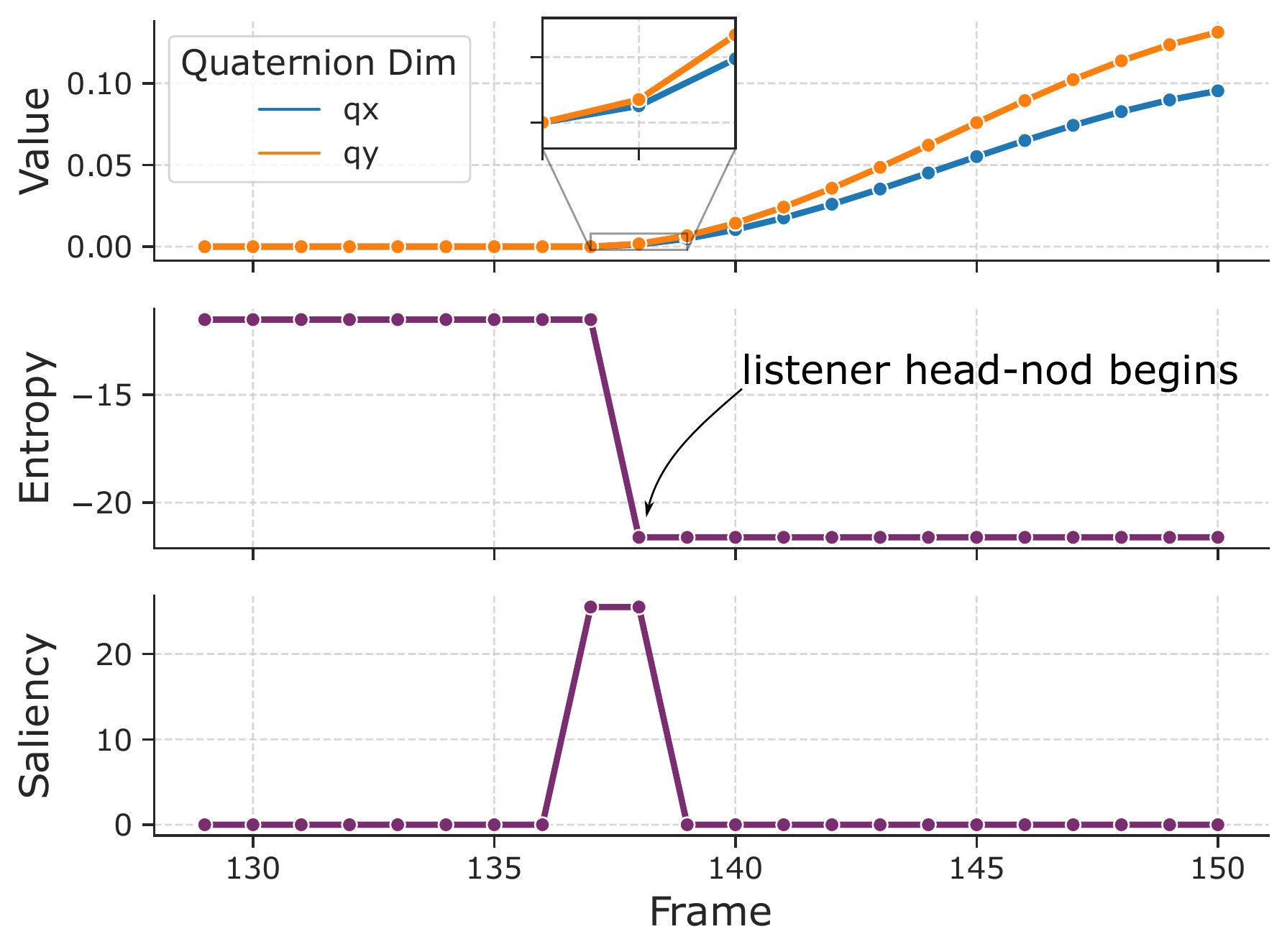}
    \caption{\textbf{Computing Saliency}. We plot the quaternion dimensions \textit{qx} and \textit{qy} for the listener that nods over $\tobs^2$ in \figurename~\ref{fig:dataset} (top). The observation of the head-nod beginning in frame $138$ makes the future over frames $183-228$ ($\tfut$) certain. This is reflected in the reduction in the mean entropy over future values of all participants (middle). The saliency map obtained using our framework (bottom) identifies the beginning of the head-nod in frame $138$ as the salient timestep towards forecasting the future over $\tfut$.}\label{fig:saliency}
\vspace{-10pt}
\end{wrapfigure}

\subsection{Computing Temporal Saliency}
By construction, it is impossible to guess the future with certainty by simply observing the current speaker speaking: there exist two valid examples of the future head behavior in the dataset for such a sequence. Only the commencement of a head-nod by a listener makes the future turn handover certain. \figurename~\ref{fig:dataset} illustrates this mechanism. Therefore, the observation of a head-nod is salient towards forecasting a turn change in \textit{this} dataset. Can our framework identify the head-nod as salient? 
Let us assume a perfect model. Such a model would predict the true variance over the two valid future quaternion trajectories given a stable speaker turn. This variance vanishes when the future becomes certain as soon as one of the listeners start nodding.
We model the future distribution using a Gaussian function for simplicity (setting std. to $10^{-10}$ for the single future), but a more complex distribution that predicts the appropriate change in variance would also work in practice.

We now implement Algorithm~\ref{alg:framework} as follows. We identify a window  where a turn change occurs in the data (frames $183$-$228$) and denote this $45$~frame window as the $\tfut$ of interest. While we manually identify an interesting event for illustration, such a window could also correspond to an interesting prediction by a model. We generate a set of candidate $\tobs$ by sliding a $30$~frame window over a horizon of $100$ frames prior to $\tfut$, with a stride of $1$~frame. For every observed $\tobs$, we fit a Gaussian density to the quaternion and speaking status features of all participants over the futures that can occur during $\tfut$. We then set the entropy of this Gaussian density as the feature $\phi$ for the $\tobs$. We obtain the saliency map using Equation~\ref{eq:saliency-map}, and plot the listener features, entropy, and saliency across timesteps in \figurename~\ref{fig:saliency}. We observe that our framework attributes higher saliency to the timesteps where the listener first begins the head-nod in frame $138$. Once the nod is in motion, the saliency drops as expected since the future is then already certain given the data. 

\section{Related Work: Explainable Methods for Time-Series Tasks Across Domains} 
\label{sec:related-work}

The larger focus of explainability techniques involving time-series data has been on the task of classifying time-series. The goal has been to attribute relevance, to each input feature at a given timestep. Here saliency approaches often overlap with techniques developed for image data, and can be categorized into gradient-based, perturbation-based, and attention-based techniques. The broad approach of gradient-based techniques involves evaluating the gradient of the output class with respect to the input \cite{baehrensHowExplainIndividual}. Several variants of this idea have been proposed \cite{sundararajan2017axiomatic, smilkov2017smoothgrad, shrikumar2017learning, lundberg2017unified, bargalExcitationBackpropRNNs2018}. The idea behind perturbation-based techniques is to examine how the output changes in response to some perturbation of the input. Here, perturbations can be implemented by either occluding contiguous regions of the input \cite{zeiler2014visualizing, mujkanovicTimeXplainFrameworkExplaining2020}; performing an ablation of the features \cite{suresh2017clinical}; or randomly permuting features \cite{molnar2020interpretable}. \citet{ismailBenchmarkingDeepLearning2020} provide a benchmark of a subset of these techniques. Attention-based methods incorporate an attention mechanism into the model that is trained to attribute importance to previous parts of the input towards an output at each timestep. Such techniques have been especially employed for healthcare data, with early methods applying a reverse-time attention \cite{choiRETAINInterpretablePredictive2016}, and later methods applying the attention to probabilistic state-space representations \cite{alaa2019attentive}. 

Some of these broad ideas have been applied to the regression setting, to make interpretable forecasts of future values of the time-series features (see Section~\ref{sec:introduction}). \citet{limTemporalFusionTransformers2020a} proposed an attention-based architecture that employs recurrent layers for local processing and interpretable self-attention layers for capturing long-term dependencies. \citet{panTwoBirdsOne2021} recently proposed a computing saliency as a mixup strategy with a learnable mask between series images and their perturbed version. They view saliency in terms of minimizing the mean squared error between the predictions and ground-truths for a particular instance. Focusing on the univariate point-forecasting problem, \citet{oreshkinNBEATSNeuralBasis2020a} proposed injecting inductive biases by computing the forecast as a combination of a trend and seasonality model. They argue that this decomposition make the outputs more interpretable. 

Developing explainable techniques for the probabilistic forecasting setting remains largely unexplored. \citet{rugamerProbabilisticTimeSeries2022} transform the forecast using an additive predictor using predefined basis functions such as Bernstein polynomials. They relate interpretability to the coefficients of these basis functions (a notion similar to that of \citet{oreshkinNBEATSNeuralBasis2020a}).
\citet{panjaInterpretableProbabilisticAutoregressive2022} embed the classical linear ARIMA model into a non-linear autoregressive neural network for univariate probabilistic forecasting. As before, the explainability here also stems from the `white-box' nature of the linear ARIMA component. \citet{liLearningInterpretableDeep2021} propose an automatic relevance determination network to identify useful exogenous variables (i.e. variables that can affect the forecast without being a part of the time-series data). To the best of our knowledge, saliency based methods have not yet been considered within this setting. 

\section{Discussion: Operationalizing Saliency for Forecasting and Implications for Interdisciplinary Research}
% \section{Discussion: Proposed Saliency Definition for Forecasting and Implications for Explainable Artificial Intelligence}
\label{sec:discussion}

We have proposed a framework for computing the saliency of observed sequences in probabilistic forecasting tasks, that results in a convenient closed-form expression of the saliency map for commonly used probability density functions to represent forecasts \cite{haNeuralRepresentationSketch2017, ramanSocialProcessesSelfSupervised2021, salzmannTrajectronDynamicallyFeasibleTrajectory2021, rudenko2020human}. 
% We begin this discussion by revisiting the different notions of saliency, to make the case for why our proposed framework is suitable for forecasting tasks. 
Rather than defining saliency in a top-down manner as a function of some task-specific error metric, we have started from a more fundamental conception of bottom-up, or task-agnostic, saliency. \citepos{loogInformationTheoreticPreattentive2011}'s original definition pertains to preattentive saliency, which captures what is perceived to be subconsciously informative before conscious (attentive) processing by the brain. Here, a surprising or unexpected observation is salient. For instance, in a large white image with a single black pixel, the black pixel is salient. The direct application of this concept to time-series data would involve identifying surprising task-agnostic temporal events. For instance, imagine viewing a static landscape where a bird suddenly flies in. The entry of the bird into the scene is unexpected, and therefore salient.  

When applied to forecasting tasks, however, this idea of surprisal (or unexpectedness or informativeness) that saliency represents needs to be tied to the future outcome. The saliency computed by most methods working on point-forecasting tasks deals with which past features are surprising given a specific realization of the future. While not explicitly stated by these works, we argue that this notion of saliency is related to the surprisal in $p_{X|Y}$ for some specific $\bm{Y}$. We therefore interpret these methods as being associative in nature within \citepos{millerExplanationArtificialIntelligence2019} categorization in Section~\ref{sec:introduction}. In contrast, our approach is counterfactual because we examine alternate future outcomes, while conceptualizing saliency more naturally defined in terms of the changes in the uncertainty in $p_{Y|X}$ in response to different realizations of observed sequences. However, rather than corresponding to random occlusions or perturbations of the input, the different realizations of $X$ in our framework correspond to real features (or behaviors) preceding a future, which is more suitable to present to domain experts as candidate causes.   

However, are we really capturing the salient predictive relationships in the data? The distribution $p_{Y|X}$ is considered to capture the structural predictive relationships between different features across the entire data. In principle, when it is possible to have access to the true $p_{Y|X}$ as in Section~\ref{sec:illustration}, the salient sequences identified by our framework reflect the \textit{true} predictive relationships between the behaviors contained in $\tobs$ and $\tfut$. However, estimating this density analytically would entail identifying the multiple futures in the data corresponding to every occurrence of the same observed sequence of features. In practice, subtle variations in behaviors and sensor measurement errors make it infeasible to estimate $p_{Y|X}$ analytically, so a model is trained to learn generalized patterns from the given data. In these cases, our framework identifies the sequences that \textit{the model considers salient} for its forecasts \textit{given the data}. Consequently, despite being grounded in a conceptually sound definition of saliency, the measures obtained by our framework may not reflect real-world truths about behavioral patterns. 
% Consequently, hypotheses about predictive behavioral insights founded on these results may not hold factu  
This is especially crucial to consider in the healthcare and human behavior domains to avoid potential prejudices against certain behaviors, or worse, misdiagnoses of affective conditions. We therefore reiterate  that the salient sequences retrieved by our framework ought to be treated as \textit{candidate} causes until subsequently examined along with a domain-expert in the loop.

% We have focused on human behavior or autonomous agent forecasting in this paper primarily due to the natural need to model a distribution over futures. However, our framework is domain agnostic and can be applied in any situation where a model outputs a probability density over the future. 

\begin{ack}
This research was partially funded by the Netherlands Organization for Scientific Research (NWO) under the MINGLE project number 639.022.606. Chirag would like to thank Amelia Villegas-Morcillo and Yeshwanth Napolean for their input on the manuscript, and the members of the TUDelft Pattern Recognition Lab for the thoughtful discussions.
\end{ack}

{
\small
\bibliography{references}
}

\end{document}